\documentclass[10pt,twocolumn,letterpaper]{article}

\usepackage{cvpr}
\usepackage{times}
\usepackage{graphicx}
\usepackage{amsmath}
\usepackage{amssymb}

\makeatletter
\@namedef{ver@everyshi.sty}{}
\makeatother
\usepackage{booktabs}
\usepackage{adjustbox}
\usepackage{cite} 
\usepackage{makecell}
\usepackage{xcolor}
\usepackage{colortbl}
\usepackage{dashrule}
\usepackage{mathtools}
\usepackage{subfigure}
\usepackage{float}
\usepackage{enumitem}
\usepackage{bm}
\usepackage{dsfont}
\usepackage{array}
\usepackage{ctable}
\usepackage{comment}
\usepackage{color}
\usepackage{lipsum}

\usepackage[hang,flushmargin]{footmisc} 

\newcommand\blfootnote[1]{%
  \begingroup
  \renewcommand\thefootnote{}\footnote{#1}%
  \addtocounter{footnote}{-1}%
  \endgroup
}

\definecolor{green1}{RGB}{93, 171, 72}
\definecolor{purple1}{RGB}{133, 109, 201}
\definecolor{orange1}{RGB}{225, 127, 14}

\DeclareMathOperator*{\argmin}{arg\,min}
\DeclareMathOperator{\vect}{vec}

\usepackage[pagebackref=true,breaklinks=true,colorlinks,bookmarks=false]{hyperref}

\cvprfinalcopy 

\def\cvprPaperID{4375} 

\ifcvprfinal\fi
\begin{document}

\title{PointNetLK Revisited}

\author{Xueqian Li$^{1}$ \quad Jhony Kaesemodel Pontes$^{1}$ \quad Simon Lucey$^{2,3}$ \\
\begin{tabular}[h]{cc}
	$^{1}$Argo AI \quad $^{2}$The University of Adelaide $^{3}$Carnegie Mellon University\\
	{\tt\small xueqianl@alumni.cmu.edu} \quad {\tt\small jpontes@argo.ai} \quad {\tt\small simon.lucey@adelaide.edu.au}
\end{tabular}
}

\maketitle
\thispagestyle{empty}

\begin{abstract}
We address the generalization ability of recent learning-based point cloud registration methods. Despite their success, these approaches tend to have poor performance when applied to mismatched conditions that are not well-represented in the training set, such as unseen object categories, different complex scenes, or unknown depth sensors. In these circumstances, it has often been better to rely on classical non-learning methods (e.g., Iterative Closest Point), which have better generalization ability. Hybrid learning methods, that use learning for predicting point correspondences and then a deterministic step for alignment, have offered some respite, but are still limited in their generalization abilities. We revisit a recent innovation---PointNetLK~\cite{aoki2019pointnetlk}---and show that the inclusion of an analytical Jacobian can exhibit remarkable generalization properties while reaping the inherent fidelity benefits of a learning framework. Our approach not only outperforms the state-of-the-art in mismatched conditions but also produces results competitive with current learning methods when operating on real-world test data close to the training set.
\end{abstract}

\section{Introduction}
\label{sc:introduction}
\begin{figure}[t!]
    \centering
    \includegraphics[width=\linewidth]{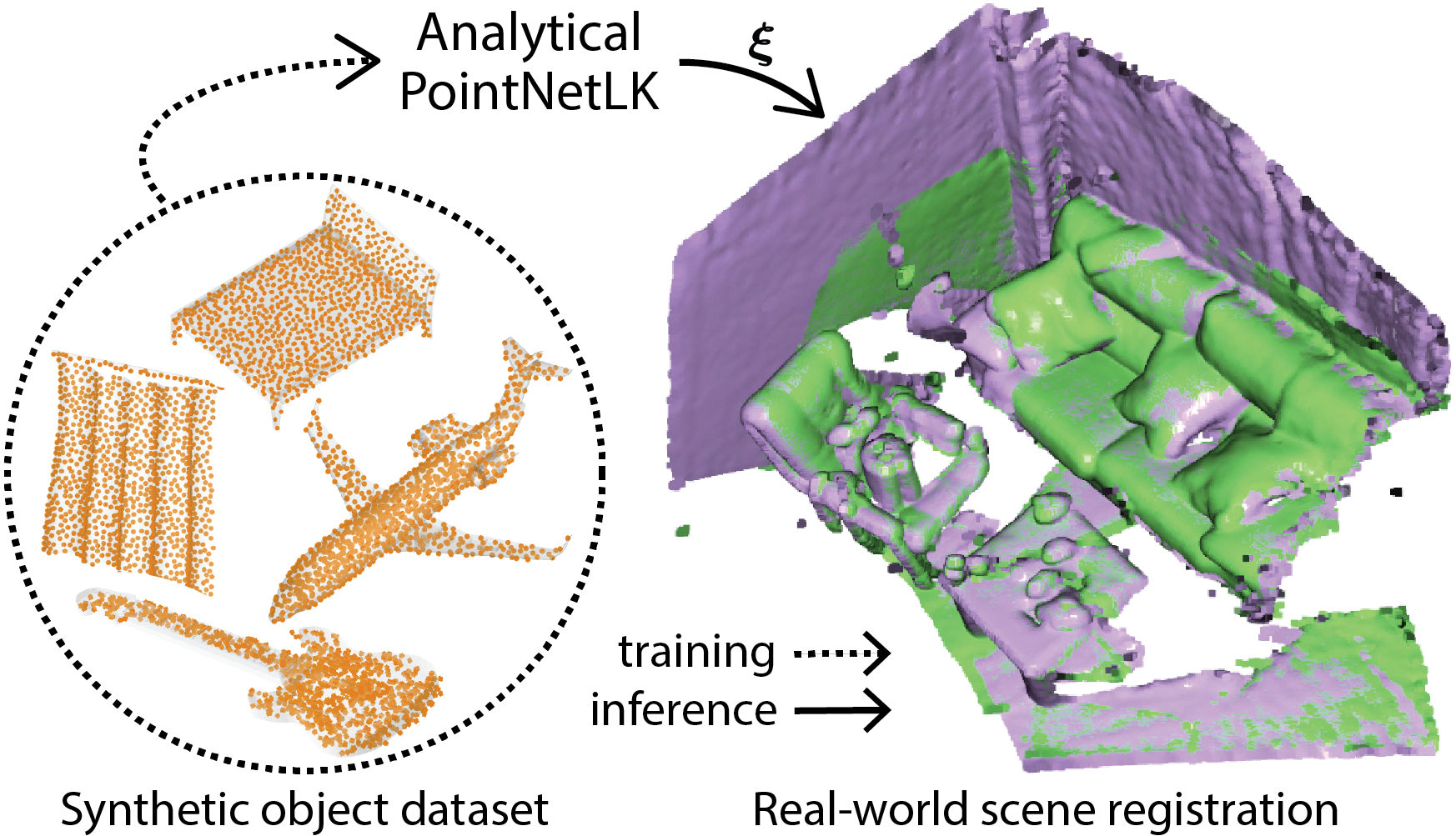}
    \caption{Our analytical derivation of PointNetLK can be trained on a clean, dense, synthetic 3D object dataset and still accurately align noisy, sparse, real-world 3D scenes. 
    \textcolor{green1}{\textbf{Green}} is the template point cloud, \textcolor{purple1}{\textbf{purple}} is the registered point cloud, and the \textcolor{orange1}{\textbf{orange}} point clouds are an object training set. $\mathbf{\xi}$ are the rigid transformation parameters inferred by our method.}
    \label{fig:teaser}
    \vspace{-4mm}
\end{figure}

Imagine a situation where you want to align two point clouds---but have little to no knowledge about the sensor noise, point density, or scene complexity before applying your algorithm (see Fig.~\ref{fig:teaser}). Under these circumstances, modern learning-based 3D point alignment methods exhibit surprisingly poor performance. This is a common experience within many industrial robotic and vision applications. The usual strategy is to instead rely on classical non-learning methods for 3D point alignment---such as Iterative Closest Point (ICP)~\cite{besl1992method}---whose performance is inferior to what is now possible through learning-based methods~\cite{sarode2019pcrnet, gross2019alignnet} but can generalize well to unknown conditions. An obvious drawback here is the hand-crafted guesswork associated with these non-learning methods, which can make their performance less than desirable.\blfootnote{Code available at \href{https://github.com/Lilac-Lee/PointNetLK_Revisited.git}{https://github.com/Lilac-Lee/PointNetLK\_Revisited.git}.}

In this paper, we advocate for a revisiting of the recent PointNetLK algorithm~\cite{aoki2019pointnetlk}. We propose a novel derivation, which circumvents many of its current limitations in terms of robustness and fidelity. Our approach employs an analytical Jacobian matrix that can be decomposed into \emph{feature} and \emph{warp} components. A specific advantage of this decomposition is the ability to modify the analytical warp Jacobian without having to re-train the PointNet features. We demonstrate how this property can be used to great effect through a voxelization process that can accommodate the alignment of complex real-world scenes (\ie, 3DMatch~\cite{zeng20173dmatch}) by training solely on clean, dense synthetic objects (\ie, ModelNet~\cite{wu20153d}). Further, in matched train-test conditions, we show that our approach outperforms state-of-the-art hybrid methods, including correspondence-based methods~\cite{wang2019deep, yuan2020deepgmr}, and the original PointNetLK of Aoki~\etal. Finally, we demonstrate how our approach can achieve results competitive with those of the state-of-the-art Deep Global Registration (DGR)~\cite{choy2020deep} method on real-world scenes, even when the latter is trained under matched conditions. These results are in stark contrast to those presented by Aoki \etal whose experimental results were limited to synthetic objects.

\noindent \textbf{Why revisit PointNetLK?}
Methods that use learning for predicting point correspondences, 
followed by a deterministic step (\eg, singular value decomposition (SVD)~\cite{wang2019deep}) to determine the alignment have become particularly popular. These \emph{hybrid methods}~\cite{wang2019deep, wang2019prnet, yew2020rpm, yuan2020deepgmr, choy2020deep}---so called because they combine modern learning with classical algorithms---exhibit some useful generalization properties. Aoki~\etal's PointNetLK approach~\cite{aoki2019pointnetlk} can also be considered a hybrid method as it leverages insights from the classical Lucas \& Kanade (LK)~\cite{lucas1981iterative} algorithm. Instead of using a neural network for modeling the entire registration pipeline, the approach learns only a point cloud embedding (\ie, PointNet~\cite{qi2017pointnet}). PointNetLK is unique to other recent hybrid methods in that it aligns directly with a learned feature representation---as opposed to using learning for predicting point correspondences. We believe it is this inherent feature-centric abstraction of the point cloud that lies at the heart of LK's generalization properties. Our analytical revisiting of the PointNetLK framework shines a clearer light upon these remarkable properties for real-world problems.

\section{Related Work}\label{sc:related}

\noindent \textbf{Classical methods:} 
The vast majority of the traditional point cloud registration methods have focused on the correspondence-based method: finding a set of correspondence candidates and using the robust algorithms to compute the registration from the noisy correspondence candidates. 
ICP, RANSAC, and their variants~\cite{zhang1994iterative, chetverikov2002trimmed, greenspan2003approximate, sandhu2008particle, yang2013go} have dominated this field for decades. 
To find reliable correspondences, various hand-crafted 3D descriptors~\cite{rusu2009fast, salti2014shot} have been proposed.
In contrast to recent deep learning methods, the classical methods tend to generalize better in unseen environments due to their lack of bias towards any pre-existing training set, making them the choice of many real-world applications nowadays.

\noindent \textbf{Hybrid learning methods:} 
Recent works have shown that the deeply learned 3D descriptors and detectors, in contrast to traditional hand-crafted counterparts, lead to less noisy correspondences and thus improve the performance when used together with ICP and RANSAC~\cite{yew20183dfeat, deng2018ppfnet, deng2018ppf, gojcic2019perfect, deng20193d, lu2019deepvcp, bai2020d3feat}. 
We refer to this type of work as a ``hybrid method'' as some components are fixed and not learned.

One notable hybrid learning work is Deep Closest Point (DCP)~\cite{wang2019deep} proposed by Wang and Solomon. 
Like other correspondence-based approaches, DCP predicts per-point descriptors and soft point correspondences, and then uses a non-learning module---SVD in the case of DCP---to estimate the final alignment. 
PRNet~\cite{wang2019prnet} further extends DCP's ability to handle partial registrations. While RPM-Net~\cite{yew2020rpm} combines PointNet with robust point matching, but requires extra point normal information.

Recently, Yuan~\etal proposed DeepGMR~\cite{yuan2020deepgmr} to learn the point correspondences through a latent Gaussian mixture model while solving registration using an efficient SVD formulation. Another approach, DGR~\cite{choy2020deep}, learns the point correspondences and uses weighted Procrustes analysis to solve for alignment. 
A fundamental issue with all these correspondence-based hybrid methods, however, is that they rely on point correspondences which might produce low-fidelity results when distinctive geometric features are scarce or unavailable.

In situations where the initial pose is unavailable, several global solutions have been applied in the literature to non-learnable~\cite{yang2013go, mellado2014super, zhou2016fast, yang2020teaser} and hybrid models~\cite{choy2020deep, jaremo2020registration}. 
In this paper, we focus on local registration problems where noisy initialization is available since in most applications the initialization can be easily obtained from previous localization results or GPS.

\noindent \textbf{Lucas \& Kanade algorithm:}\label{sc:deeplk}
Recent works have shown promising results when applying the Lucas \& Kanade (LK) algorithm~\cite{lucas1981iterative}, originally used in image alignment, to the point cloud registration problem. The LK algorithm and its variants~\cite{bouguet2001pyramidal, baker2004lucas, lucey2012fourier, oron2014extended, lin2016conditional} seek to minimize the alignment error between two images by using either extracted distinct features or all the pixels in an image (\ie, photometric error). Lv~\etal~\cite{lv2019taking} used a neural network to extract pyramid features for image tracking. Wang~\etal~\cite{wang2018deep} proposed a regression-based object tracking framework, which uses the LK algorithm in an end-to-end deep learning pipeline to train image feature descriptors. 
In PointNetLK~\cite{aoki2019pointnetlk}, the authors expanded the end-to-end LK tracking paradigm to the 3D point cloud.

\section{Background}
\label{background}
\noindent \textbf{Problem statement:}
Let $\mathbf{P}_\mathcal{T}\;{\in}\;\mathbb{R}^{N_1\times3}$ and  $\mathbf{P}_\mathcal{S}\;{\in}\;\mathbb{R}^{N_2\times3}$ be the template and source point clouds respectively, where $N_1$ and $N_2$ are the number of points. The rigid transformation that aligns the observed source $\mathbf{P}_\mathcal{S}$ to the template $\mathbf{P}_\mathcal{T}$ can be defined as $\mathcal{G}(\bm{\xi}) {=} \exp(\sum_{p=1}^{6} \xi_{p}\mathbf{T}_{p}) \;{\in}\; \mbox{SE}(3)$, where $\bm{\xi} \;{\in}\; \mathbb{R}^{6}$ are the exponential map twist parameters, $\mathbf{T}$ are the generator matrices, and SE(3) is the special Euclidean group. The PointNet embedding function $\phi:\mathbb{R}^{N\times3} \rightarrow \mathbb{R}^{K}$ can be employed to encode a 3D point cloud into a fixed $K$-dimensional feature descriptor. 
Thus, the point cloud registration optimization can be formulated as finding $\bm{\xi}$ that minimizes the feature difference between the source point cloud and the template point cloud as
\begin{align}
    \argmin_{\bm{\xi}} \lVert \phi \left(\mathcal{G} \left(\bm{\xi} \right) {\cdot} \mathbf{P}_\mathcal{S} \right) - \phi\left(\mathbf{P}_\mathcal{T}\right) \rVert^2_2,
    \label{eq_main_objective}
\end{align}
where the operator $(\cdot)$ denotes the rigid transformation.

\noindent \textbf{PointNetLK:}
Given the non-linearity of Eq.~\eqref{eq_main_objective}, PointNetLK iteratively optimizes for an incremental twist parameter $\Delta\bm{\xi}$---as in the classical LK algorithm---that best aligns the point cloud feature embeddings $\phi$ learned by PointNet. For efficiency, the twist increments $\Delta \bm{\xi}$ are applied to $\mathbf{P}_\mathcal{T}$ instead of $\mathbf{P}_\mathcal{S}$, as in the inverse compositional formulation of the LK algorithm (IC-LK)~\cite{baker2004lucas}. Thus, the warp increment $\Delta \bm{\xi}$ is obtained by linearizing
\begin{align}
    \argmin_{\Delta\bm{\xi}} \lVert \phi \left(\mathbf{P}_\mathcal{S} \right) {-} \phi \left(\mathcal{G}^{\text{-}1} \left( \bm{\xi}^i {\circ^{\text{-}1}} \Delta \bm{\xi} \right) {\cdot} \mathbf{P}_\mathcal{T} \right) \rVert^2_2,
    \label{eq_main_objective_incremental_warp}
\end{align}
using first-order Taylor expansion. The symbol ${\circ^{\text{-}1}}$ refers to the inverse composition. This produces
\begin{align}
    \argmin_{\Delta\bm{\xi}} \lVert \phi {\left(\mathbf{P}_\mathcal{S} \right) {-} \phi \left(\mathcal{G}^{\text{-}1} \left(\bm{\xi}^i \right) {\cdot} \mathbf{P}_\mathcal{T} \right)} {-} \mathbf{J} \Delta \bm{\xi} \rVert^2_2,
    \label{eq_main_objective_incremental_warp_linear}
\end{align}
which is then solved for $\eta$ iterations as $\bm{\xi}^{(i+1)} \leftarrow \bm{\xi}^i \circ^{\text{-}1} \Delta\bm{\xi}$.
The Jacobian matrix $\mathbf{J}$ is defined as
\begin{align}
    \mathbf{J} & = \frac{\partial \phi \left(\mathcal{G}^{\text{-}1} \left(\bm{\xi} \right) {\cdot} \mathbf{P}_\mathcal{T} \right) }{\partial \bm{\xi}^T} \: \in \mathbb{R}^{K\times6}, \label{eq_jac}
\end{align}
where each twist parameter gradient $\mathbf{J}_p \in \mathbb{R}^{K}$ is approximated using numerical finite differences as
\begin{align}
    \mathbf{J}_p \approx \frac{\phi \left( \exp(-t_p \mathbf{T}_{p}) {\cdot} \mathbf{P}_{\mathcal{T}}\right) - \phi(\mathbf{P}_{\mathcal{T}})}{t_p},
    \label{eq_stochastic}
\end{align}
where $t_p$ is the step size to infinitesimally perturb the twist parameter $\xi_p$ of $\mathbf{J}_p$. Instead of learning $t_p$, PointNetLK requires a pre-defined value for the approximation, and in practice, is set to be the same for each dimension, \ie, $t_p{=}t,~\forall p$.\footnote{Refer to IC-LK~\cite{baker2004lucas} and PointNetLK~\cite{aoki2019pointnetlk} for in-depth information.}

\noindent \textbf{Limitations:}
PointNetLK draws inspiration from the classical IC-LK in which both seek to align a source to a template (image or point cloud) by minimizing the sum of squared error between their features through gradient descent. 
Also, both methods switch the role of the template and source to avoid the re-evaluation of the Jacobian in every iteration and solve for an incremental warp that is inverted before composed to the current estimate. 
However, a fundamental difference between PointNetLK and IC-LK is in the derivation of the Jacobian matrix. The derivation in IC-LK assumes a particular form for the Jacobian---it is defined as a product of two gradients: an image/feature gradient and a warp Jacobian. Such a Jacobian decoupling is useful as the features and the warp parameters can be separated from each other. This allows for an easy replacement of the warp function according to the registration task. 
PointNetLK overlooked this important property and proposed a derivation based on a numerical finite difference technique which is numerically unstable and can result in issues such as biased gradient values (when a very large step size $t$ is selected), or catastrophic cancellation (when a very small step size $t$ is selected). 
In addition to the above numerical issues, PointNetLK computes the gradients w.r.t. $p$ parameters, 
which might easily become a computational overhead during training.

\section{Analytical PointNetLK}
\label{deterministic_ptnetlk}
We propose an analytical derivation of PointNetLK that circumvents its current limitations.
Instead of approximating the Jacobian using numerical finite differences, we propose to decompose the Jacobian in Eq.~\eqref{eq_jac} using the chain rule into two components: \emph{feature gradient} and \emph{warp Jacobian}:
\begin{align}
    \mathbf{J} = \underbrace{\frac{\partial \phi \left(\mathcal{G}^{\text{-}1} \left( \bm{\xi} \right) \cdot \mathbf{P}_\mathcal{T} \right) }{\partial \left(\mathcal{G}^{\text{-}1} \left( \bm{\xi} \right) \cdot \mathbf{P_\mathcal{T}} \right)^T}}_\textrm{Feature gradient} ~ \underbrace{\vphantom{ \frac{a}{\left(\bm{\xi}\mathbf{P_\mathcal{T}}\right)^T}} \frac{\partial \left(\mathcal{G}^{\text{-}1} \left( \bm{\xi} \right) \cdot \mathbf{P}_\mathcal{T} \right)}{\partial \bm{\xi}^T}}_\textrm{Warp Jacobian}. \label{eq_general_jac}
\end{align}
Assuming without loss of generality that the initial transformation applied to the template point cloud, $\mathbf{P}_{\mathcal{T}}$, is the identity matrix, \ie, $\mathcal{G}^{\text{-}1} \left( \bm{\xi}\right){=}\mathbf{I}$, we can pre-compute $\mathbf{J}$ as
\begin{align}
    \mathbf{J} = \underbrace{\frac{\partial \phi(\mathbf{P}_\mathcal{T}) }{\partial \vect(\mathbf{P_\mathcal{T}})^T} \vphantom{\frac{\partial(\mathcal{G}^{\text{-}1} \left( \bm{\xi} \right) \cdot \mathbf{P}_\mathcal{T})}{\partial \bm{\xi}^T}}}_\textrm{Feature gradient} ~ \underbrace{ \frac{\partial(\mathcal{G}^{\text{-}1} \left( \bm{\xi} \right) \cdot \mathbf{P}_\mathcal{T})}{\partial \bm{\xi}^T}}_\textrm{Warp Jacobian}, \label{eq_simplified_jac}
\end{align}
where $\vect(\cdot)$ is the vectorization operator. The Jacobian is therefore a constant and its re-computation in each iteration can be avoided.
The feature gradient $\frac{\partial\phi(\mathbf{P}_\mathcal{T})} {\partial \vect(\mathbf{P_\mathcal{T}})^T} \in \mathbb{R}^{N_1\times 3 \times K}$ describes how the PointNet feature embedding changes w.r.t the template point cloud $\mathbf{P_\mathcal{T}}$.
The warp Jacobian $\frac{\partial(\mathcal{G}^{\text{-}1} \left( \bm{\xi} \right) \cdot \mathbf{P}_\mathcal{T})}{\partial \bm{\xi}^T} \in \mathbb{R}^{N_1\times3\times 6}$ describes how changes in the twist parameters $\bm{\xi}$ affect the rigid transformation of $\mathbf{P_\mathcal{T}}$.

We employ a simplified PointNet architecture~\cite{qi2017pointnet} (\ie, with $L{=}3$ layers and without the T-Net module) to extract per-point features $\mathbf{z}_{l}$ from $\mathbf{P}_{\mathcal{T}}$. The feed-forward network is defined as $\mathbf{z}_{l} {=} \mbox{ReLU} (\mbox{BN}_{l}(\mathbf{A}_{l} \mathbf{z}_{l-1} + \mathbf{b}_{l}))$,
where $\mathbf{A}$ is a matrix transformation, $\mathbf{b}$ is the bias term, $\mbox{BN}(\cdot)$ is the batch normalization layer, $\mbox{ReLU}(\cdot)$ is the element-wise rectified linear unit function, and $l$ is the network's $l$-th layer. 
Thus, our per-point embedding feature can be simplified as $\mathbf{z}_L$.
We explicitly solve for the partial derivatives of the embedding feature $\mathbf{z}_L$ w.r.t the 
input template point cloud $\mathbf{P}_{\mathcal{T}}$ as
\begin{align}
    \frac{\partial \mathbf{z}_L}{\partial \vect(\mathbf{P}_{\mathcal{T}})^T} = \prod^{L}_{l=1} \frac{\partial \mathbf{z}_l}{\partial \mathbf{z}_{l-1}^{T}},
    \label{eq_feat_jac}
\end{align}
where $\mathbf{z}_0 {=} \vect(\mathbf{P}_{\mathcal{T}})$. 
To extract a global feature vector, we apply the max pooling operation, $\mbox{Pool}(\cdot)$, since it is a symmetrical function that is invariant to the unordered nature of point clouds (as in~\cite{qi2017pointnet, aoki2019pointnetlk}). Thus, the final Jacobian is defined as
\begin{align}
    \mathbf{J} = \mbox{Pool} \left(\underbrace{\frac{\partial \mathbf{z}_L}{\partial \vect(\mathbf{P}_{\mathcal{T}})^T}}_\textrm{Feature gradient} ~ \underbrace{\vphantom{\frac{\partial \mathbf{z}_L}{\partial \vect(\mathbf{P}_{\mathcal{T}})^T}} \frac{\partial \left(\mathcal{G}^{\text{-}1} \left( \bm{\xi}\right) \cdot \mathbf{P}_\mathcal{T} \right)}{\partial \bm{\xi}^T}}_\textrm{Warp Jacobian} \right). \label{eq_final_jac}
\end{align}
The transformations are then updated for $\eta$ iterations as,
$\bm{\xi}^{(i+1)} \leftarrow \bm{\xi}^i {\circ} ^{\text{-}1} \Delta \bm{\xi}$, 
$\mathcal{G}^{\text{-}1} \left( \bm{\xi} \circ^{\text{-}1} \Delta \bm{\xi} \right) = \mathcal{G}^{\text{-}1} \left( \bm{\xi} \right) \mathcal{G}^{\text{-}1} \left( \Delta \bm{\xi} \right)$.

Our proposed analytical PointNetLK does not rely on numerical finite difference techniques as in the original PointNetLK. Therefore, it does not depend on the step size parameters to approximate the gradients, which might lead to numerical instabilities (see the analysis in the subsection~\ref{sc:jacobian_analysis}). Also, our analytical Jacobian is decomposable, meaning that we can reuse the feature gradient and change the warp Jacobian component according to the registration task without retraining the entire registration pipeline (see Fig.~\ref{fig:decomposition}). This was previously not possible with the original PointNetLK formulation.

Note that our Jacobian matrix is fully analytical (for both feature gradient and warp Jacobian) in contrast to the IC-LK algorithm which only the warp Jacobian component is analytical. Our method's feature gradient is analytical since it is computed directly over the 3D geometric coordinates of a point cloud.
Our method differs from the recent image-based conditioned LK~\cite{lin2016conditional} and Deep-LK~\cite{wang2018deep} in that we directly compute the gradients over a non-linear function $\phi$ instead of a linearly regressed function.

Furthermore, we emphasize that an analytical Jacobian computation is fundamentally different from the backpropagation process commonly used in neural networks. 
Backpropagation relies on the computation of gradients of a loss function with respect to the network weights. Instead, we propose to unroll the network and explicitly compute the feature gradients with respect to the input point cloud.
Finally, we argue that automatic differentiation (AD) is not suitable to solve the analytical Jacobian because it is not decomposable. Even if we only use AD to compute the feature gradient, it still incurs a computational overhead given the high dimensionality of the PointNet features (\ie, $1024$).

\subsection{Voxelized analytical PointNetLK}
\label{sc:voxelization}
In our proposed analytical PointNetLK, we employ PointNet to extract a global feature descriptor from a point cloud.
This has been shown to work well to represent 3D synthetic objects~\cite{aoki2019pointnetlk, qi2017pointnet}. However, as we show in our experiments, such a global feature embedding is not able to capture complex, large scene features, and the performance drops drastically when trying to register real-world scenes.

Here, we propose a strategy to tackle the registration of a complex, real-world scene by first dividing it into smaller regions. 
During testing, we voxelize the 3D space and use PointNet to extract local feature descriptors from each voxel.
Thus, the feature embedding of each voxel can capture local geometric information despite the complexity of the scene.
Formally, given a point cloud $\mathbf{P}$ sampled from a complex scene, we first partition the set of points into local voxels, where the points inside a voxel are represented as $\mathbf{V}_m$, where $m{=}1{:}M$, and $M$ is the number of voxels.
Then, for each voxel, a local Jacobian $\mathbf{J}_{\mathbf{V}_m}$ is computed within its own coordinate frame. However, we still want to solve for a global transformation that aligns the entire point cloud at once.
To achieve this, we employ a conditioned warp Jacobian to transfer the local Jacobians into the global coordinate frame as
\begin{align}
    \mathbf{J}_g = \left[ \mathbf{J}_{\mathbf{V}_1}, \cdots, \mathbf{J}_{\mathbf{V}_M} \right] 
    \begin{bmatrix}\left(\frac{\partial \bm{\xi}_{\mathbf{V}_1}}{\partial \bm{\xi}^T}\right)\\
    \vdots \\
    \left(\frac{\partial \bm{\xi}_{\mathbf{V}_M}}{\partial \bm{\xi}^T}\right)
    \end{bmatrix} . \label{eq_voxel_jac_3}
\end{align}
Thus, the optimization defined in Eq.~\eqref{eq_main_objective_incremental_warp_linear} becomes,
\begin{align}
    \argmin_{\Delta\bm{\xi}} \lVert \sum_{m=1}^{M} {\left( \phi {\left(\mathbf{V}_{\mathcal{S}m} \right) {-} \phi \left(\mathcal{G}^{\text{-}1} \left(\bm{\xi}^i \right) {\cdot} \mathbf{V}_{\mathcal{T}m} \right)} \right)} {-} \mathbf{J}_g \Delta \bm{\xi} \rVert^2_2.
    \label{eq_main_objective_incremental_warp_linear_voxelization}
\end{align}
The sum over the local feature gradient from each voxel represents the global feature of the entire point cloud, which corresponds to the global analytical Jacobian.\footnote{Please refer to the supplementary material for more details regarding the voxelization strategy.}

\begin{figure}[t!]
    \centering
    \includegraphics[width=\linewidth]{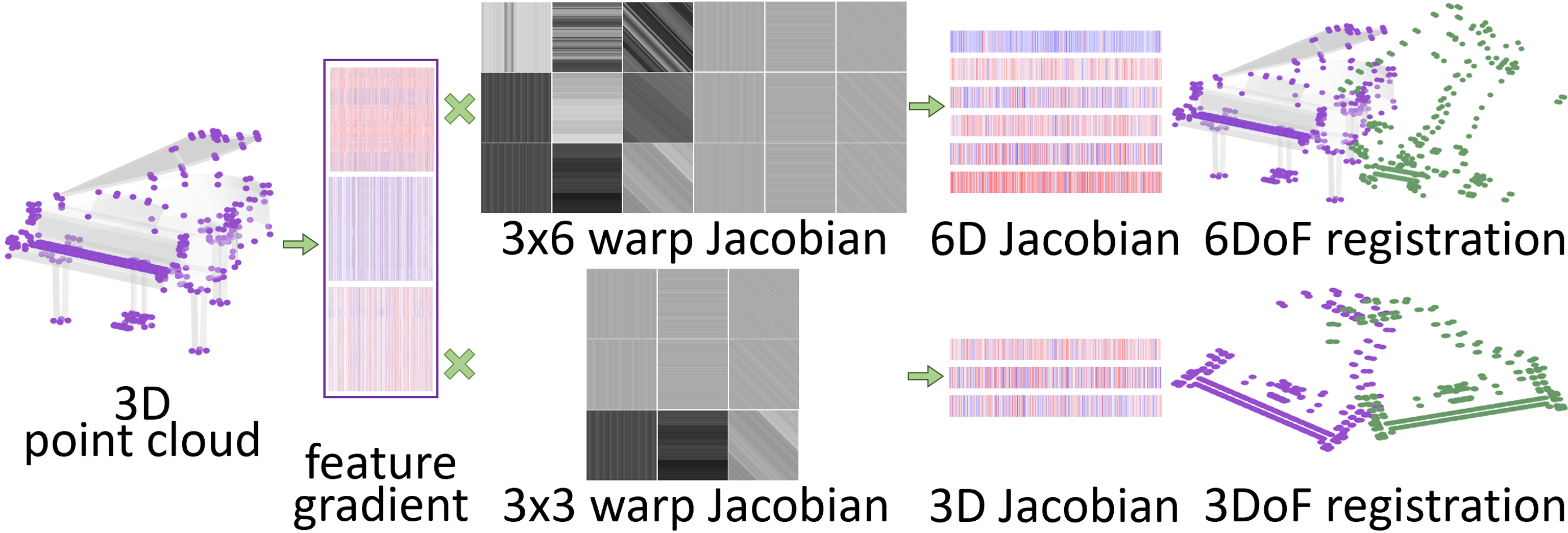}
    \caption{\textbf{Jacobian decomposition.} We can pre-compute the analytical feature gradient from a 3D point cloud. For different registration tasks, we do not need to re-train the entire registration pipeline. Only the feature gradient is learnable, with the warp Jacobian being defined analytically and easily modified. For example, in 6 DoF registration, we can compose a $3{\times}6$ warp Jacobian to the pre-computed feature gradient to get the steepest descent point features as depicted in the upper part. If we impose a constrained 3 DoF rigid transformation (\eg, the rigid transformation in the 2D $xy$-plane) as shown in the bottom part, a $3{\times}3$ warp Jacobian can be computed and composed to get a 3D Jacobian.}
    \label{fig:decomposition}
\end{figure}

\begin{figure}[t!]
    \centering
    \includegraphics[width=0.95\linewidth]{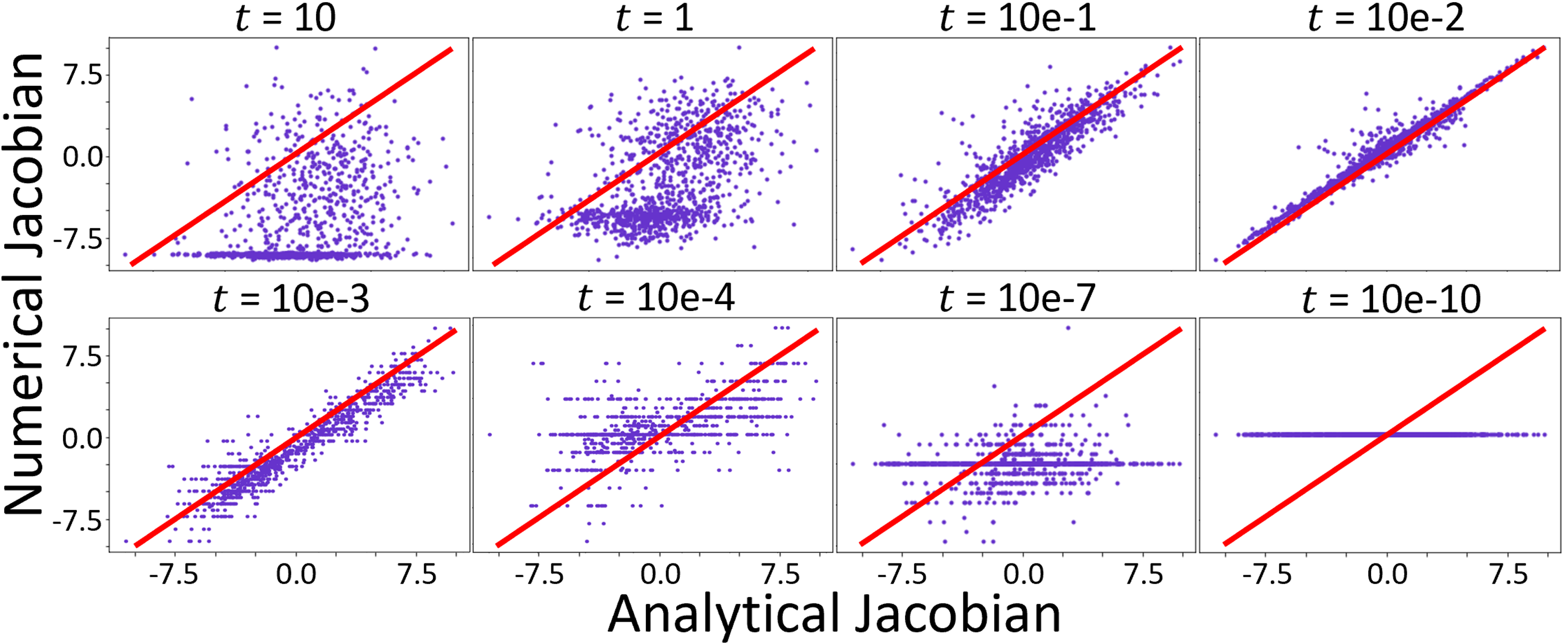}
     \caption{\textbf{Analysis of the Jacobian.} We demonstrate the correlation between the numerical and analytical Jacobian for different step sizes $t$. Given that the Jacobian matrix is controlled by 6 parameters,
     we only plot a single dimension (rotation w.r.t. z-axis) to simplify the visualization. Note how the numerical Jacobian proposed in the original PointNetLK~\cite{aoki2019pointnetlk} is extremely sensitive to different $t$'s. The diagonal red line represents the ideal error (\ie, zero error) between the numerical and analytical Jacobian, meaning that the closer the points are to the red line, the better the numerical gradient approximation is. A large (\eg, $t{\geq}1$) or a small (\eg, $t{\leq}10e{-}4$) step size drastically impacts the numerical approximation of the Jacobian.}
    \label{fig:jacobian}
    \vspace{-4mm}
\end{figure}

Our proposed voxelized formulation to handle the registration of complex scenes is only possible because we can decompose the analytical Jacobian (see Fig.~\ref{fig:decomposition}). 
We only modify the analytical warp Jacobian and re-use the feature gradients even if it was previously trained on a completely different dataset. For example, one can train on a synthetic object dataset and test on large, complex scenes.

\subsection{Analysis of the Jacobian}
\label{sc:jacobian_analysis}
We demonstrate how the numerical Jacobian deviates from the analytical Jacobian according to the step size $t$ in Fig.~\ref{fig:jacobian}. We changed $t$ from $10e{-}10$ to $10$ to show how unstable the numerical Jacobian might become.
Although the step size $t{=}10e{-}2$ approximates the Jacobians to the analytical ones, it is still noisy.
Theoretically, $t$ would need to be infinitesimally small to approximate to the analytical Jacobian using a numerical finite difference method. 
Practically, if we set $t$ to a large value (\eg, $t{=}10$), the gradients are extremely noisy and biased, while with a small $t$ (\eg, $t{=}10e{-}10$) the gradients vanish. 
Moreover, the step size is highly dependent on the dataset, which might not always be easily tuned.

Note that heuristic techniques such as gradient clipping will likely not avoid numerical instabilities
since the inherent problem of numerical finite differences cannot be solved.
The analytical Jacobian circumvents numerical limitations and allows for high fidelity registration, as shown in the experiments in Section~\ref{ex:fidelity}.

\subsection{Loss function}
\label{loss_function}
We use two loss functions to train our proposed analytical PointNetLK: a rigid transformation loss and a feature loss.

\noindent \textbf{Rigid transformation loss:}
We want to minimize the mean squared error (MSE) between the estimated transformations $\Tilde{\mathcal{G}}$ and the ground-truth transformations $\mathcal{G}$. For efficiency, we formulate the transformation loss as
\begin{align}
    \mathcal{L}_{\mathcal{G}} = \min_{\Delta \bm{\xi}} \lVert \Tilde{\mathcal{G}}\left( \bm{\xi} \circ \Delta \bm{\xi} \right) \mathcal{G}^{\text{-}1}\left( \bm{\xi}\right) - \mathbf{I}_4 \rVert_F^2,
    \label{loss_t}
\end{align}
where $\mathbf{I}_4 \;{\in}\; \mathbb{R}^{4 \times 4} $ is an identity matrix, and $\|\cdot\|_F$ is the Frobenius norm. This formulation is computationally efficient as it avoids matrix logarithm operations.

\noindent \textbf{Feature loss:}
To capture different feature cues for the transformed point clouds, we include a feature loss during training. We want to minimize the error between the template $\phi(\mathbf{P}_{\mathcal{T}})$ and the source point cloud feature $\phi(\mathbf{P}_{\mathcal{S}})$. 
When the point clouds are aligned,
the feature difference should be close to zero. Thus the feature loss is defined as
\begin{align}
    \mathcal{L}_{\phi} {=} \min_{\Delta \bm{\xi}} \lVert \phi \left(\Tilde{\mathcal{G}}^{\text{-}1} \left(\bm{\xi} \circ^{\text{-}1} \Delta \bm{\xi} \right) {\cdot} \mathbf{P}_\mathcal{T} \right) {-} \phi \left( \mathbf{P}_\mathcal{S} \right)\rVert^2_2.
    \label{loss_phi}
\end{align}

\section{Experiments}
\label{experiments}
We evaluated our analytical PointNetLK on both synthetic and real-world datasets. Our results show that our method, in addition to its high fidelity, generalized 
better than the state-of-the-art baseline 
when trained on a synthetic object dataset and tested on a real-world scene dataset.

\noindent\textbf{Datasets:}
We used the following datasets for experiments: 

\noindent\textbf{1. ModelNet40}~\cite{wu20153d} contains synthetic 3D objects from 40 categories ranging from airplane, car, to plant, lamp, etc. To demonstrate the generalizability of our method, we chose 20 categories for training and the remaining 20 for testing. We also partitioned 20\% of the training set for evaluation.

\noindent\textbf{2. ShapeNetCore}~\cite{chang2015shapenet} contains 3D shapes from 55 object categories. We chose 12 categories for testing.

\noindent\textbf{3. 3DMatch}~\cite{zeng20173dmatch} is an assembly of several real-world indoor datasets, containing complex 3D RGB-D scans obtained from multiple scenes as hotel, office, kitchen, etc. We chose 8 categories from 7-Scenes~\cite{shotton2013scene} and SUN3D~\cite{xiao2013sun3d} to generate 3D scan pairs with at least 70\% overlap for testing, while 48 categories for training, and 6 for validation. 

\noindent\textbf{Setup:}
The rigid transformations used for training and testing included rotations randomly drawn from $[0^\circ, 45^\circ]$ and translations randomly sampled from $[0, 0.8]$ for fair comparisons. We sampled $1,000$ points from the objects' vertices, and all point clouds were centered at the origin within a unit box during training. We applied the rigid transformations to the source point clouds to generate the template point clouds.
Given that some correspondence-based methods require significant computation, we only sampled up to $1,000$ points during testing if not specified. 
We set the maximum number of iterations to 10 (or 20 for 3DMatch dataset) for all the iterative methods. 
All testing experiments were performed on a single NVIDIA TITAN X (Pascal) GPU or an Intel Core i7-8850H CPU at 2.60 GHz. We adapted the code released by the other methods for our experiments.

\noindent\textbf{Metrics:}
We used the difference between the estimated transformations and the ground truth transformations as the error metric. Note that the rotation errors were measured in degrees.
The root mean squared error (\textbf{RMSE}) evaluates the variation of each error, and the median error (\textbf{Median}) captures the error distribution. 
We set the success registration criterion to be a rotation error smaller than $5^\circ$ and a translation error smaller than $0.05$ if not specified. 
We measured the ratio of successful alignments to the total number of point cloud pairs as the \textbf{success ratio}. 
The area under the curve (\textbf{AUC}) was then defined as the 2-D area under the success ratio curve.

\noindent\textbf{Implementation details:}
We trained DCP~\cite{wang2019deep}, DeepGMR~\cite{yuan2020deepgmr}, PointNetLK~\cite{aoki2019pointnetlk} and our method on the synthetic ModelNet40 dataset if not specified. For the analytical PointNetLK, we used the Adam optimizer~\cite{kingma2014adam} with learning rate of $1e\text{-}3$, and a decay rate of $1e\text{-}4$.

The PRNet~\cite{wang2019prnet} method is a successful hybrid learning method. However, we were not able to train it properly to get convincing results due to random indexing errors and large training losses (the same pattern observed in~\cite{choy2020deep}). 
As mentioned in other works~\cite{yuan2020deepgmr, dang2020learning, li2020unsupervised}, PRNet seems to achieve better results than DCP~\cite{wang2019deep}, but not drastically. Moreover, the FMR~\cite{huang2020feature} method is a derivative of the original PointNetLK~\cite{aoki2019pointnetlk} with a different loss function. Therefore, we reported its performance in our ablation study.

\begin{table}[t!]
\begin{adjustbox}{width=\columnwidth}
\centering
\begin{tabular}{lcccccccc}\toprule
                & \multicolumn{4}{c}{ModelNet40} & \multicolumn{4}{c}{ShapeNetCore} \\ \cmidrule(lr){2-5} \cmidrule(lr){6-9}
                 & \multicolumn{2}{c}{Rot. Err. (deg.)~$\downarrow$} & \multicolumn{2}{c}{Trans. Err.~$\downarrow$} & \multicolumn{2}{c}{Rot. Err. (deg.)~$\downarrow$} & \multicolumn{2}{c}{Trans. Err.~$\downarrow$} \\ 
                 \cmidrule(lr){2-3} \cmidrule(lr){4-5} \cmidrule(lr){6-7} \cmidrule(lr){8-9}
                Algorithm & RMSE & Median & RMSE & Median & RMSE & Median & RMSE & Median \\ \midrule
ICP~\cite{besl1992method} & 39.33 & 5.036 & 0.474 & 0.058 & 40.71 & 5.825 & 0.478 & 0.073 \\
DCP~\cite{wang2019deep} & 5.500 & 1.202 & 0.022 & 0.004 & 8.587 & 0.930 & 0.021 & 0.003 \\
DeepGMR~\cite{yuan2020deepgmr} & 6.059 & 0.070 & \textbf{0.014} & 8.42e-5 & 6.043 & 0.013 & \textbf{0.005} & 9.33e-6 \\
PointNetLK~\cite{aoki2019pointnetlk} & 8.183 & 3.63e-6 & 0.074 & 5.96e-8 & 12.941 & 4.33e-6 & 0.115 & 5.96e-8 \\
Ours   & \textbf{3.350} & \textbf{2.17e-6} & 0.031 & \textbf{4.47e-8} & \textbf{3.983} & \textbf{2.06e-6} & 0.049 & \textbf{2.98e-8} \\ \bottomrule
\end{tabular}
\end{adjustbox}
\vspace{0.01ex}
\caption{\textbf{Accuracy and generalizability.} Results on unseen object categories from ModelNet40 and ShapeNetCore. Our method outperformed other methods in most metrics showing its accuracy and generalizability.
$\downarrow$ means smaller values are better.}
\label{tb:accuracy}
\end{table}

\subsection{Performance on synthetic datasets}
\label{ex:synthetic}
\begin{figure}[t!]
\centering
    \begin{subfigure}
        \centering
            \includegraphics[width=0.489\linewidth]{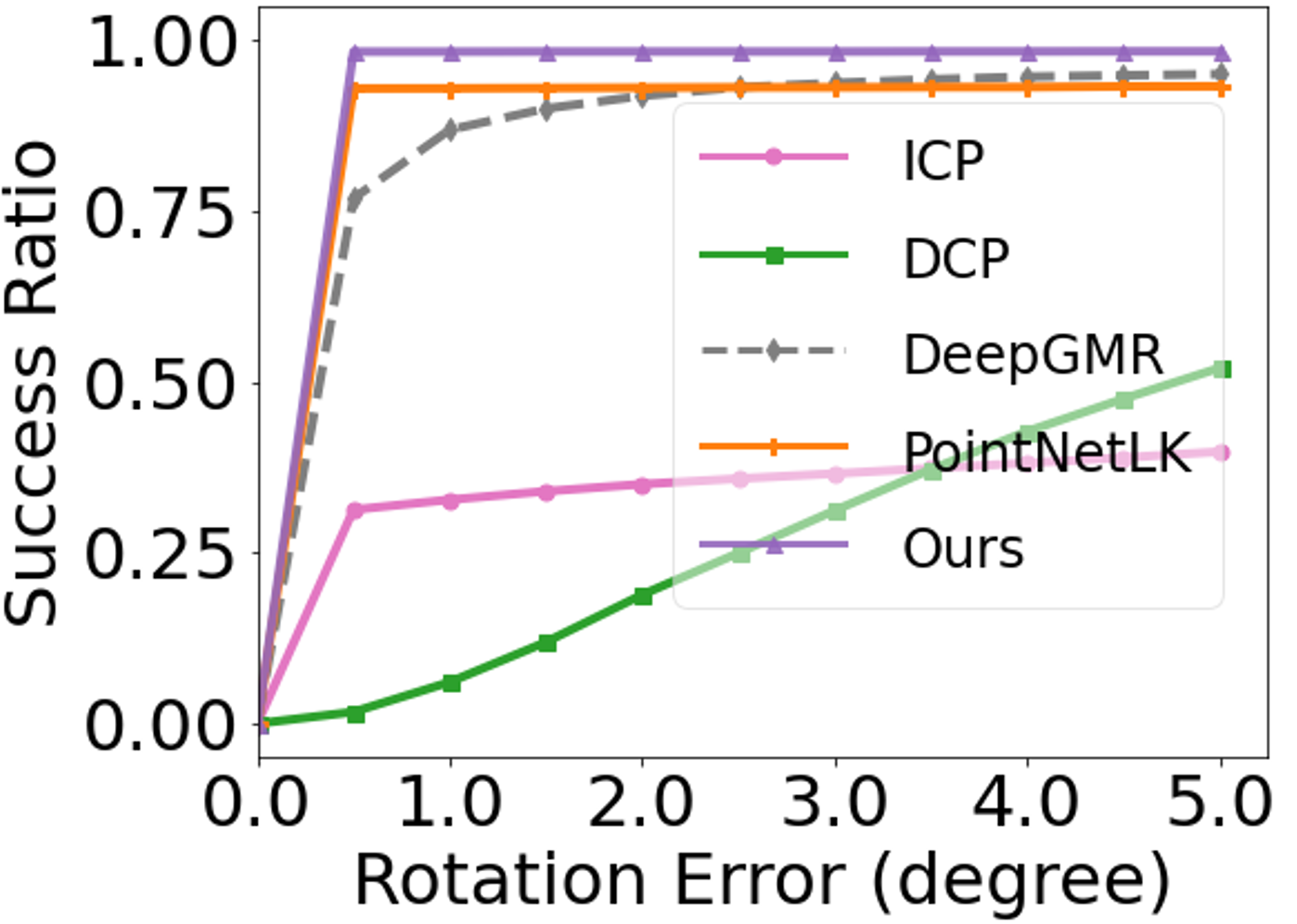}
        \label{fig:acc_rot_01}
    \end{subfigure}
    \begin{subfigure}
        \centering
         \includegraphics[width=0.489\linewidth]{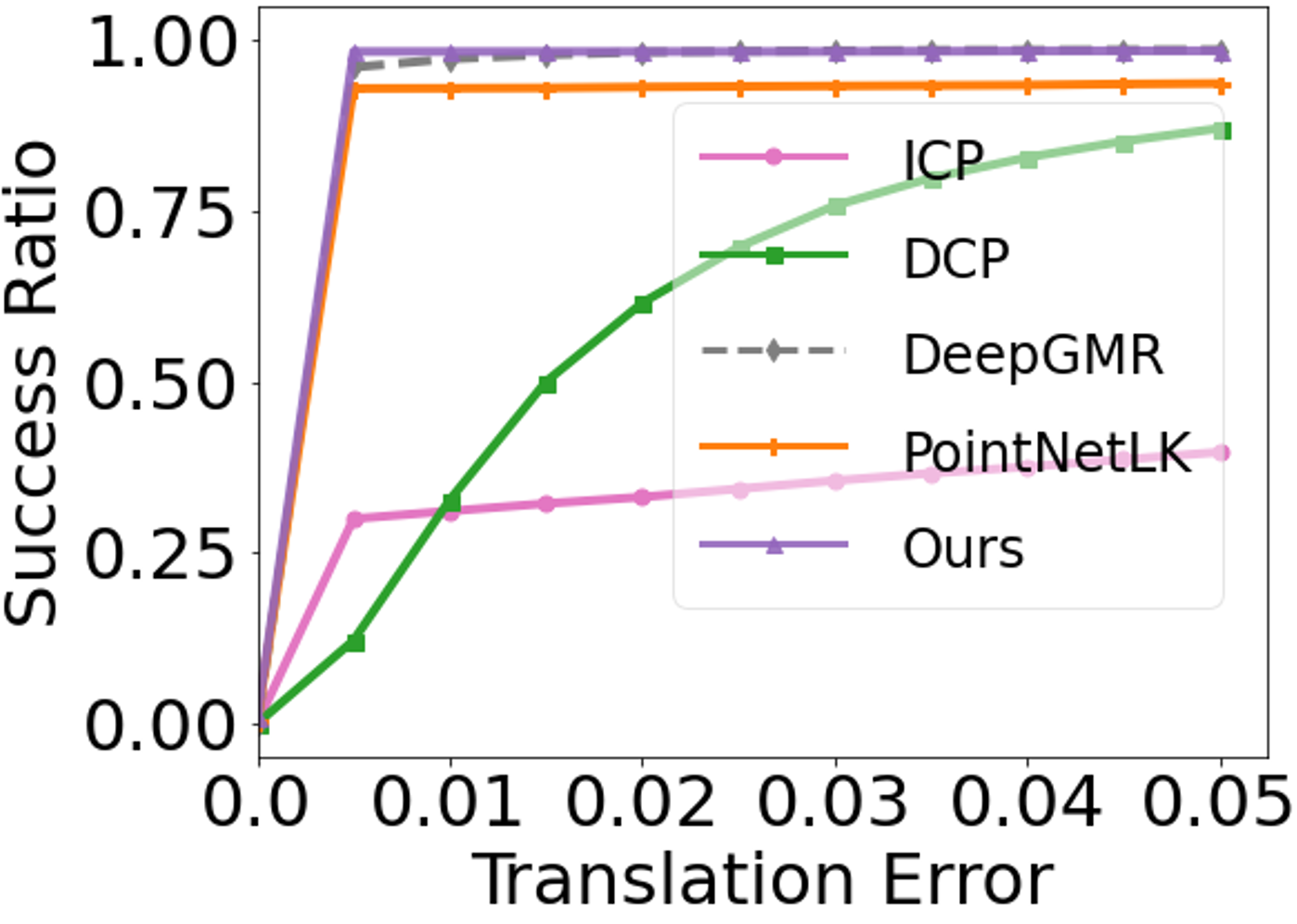}
         \label{fig:acc_trans_01}
    \end{subfigure}
    \caption{\textbf{Accuracy.} 
    The purple line shows that our method achieved nearly $100\%$ of success for alignments with a small maximum error threshold which indicates an advantage over other methods.}
    \label{fig:accuracy}
\end{figure}

\noindent\textbf{Accuracy and generalization across object categories:}
\label{ex:accuracy}
We report our method's accuracy compared against other methods in Fig.~\ref{fig:accuracy} and Table~\ref{tb:accuracy}. 
As shown in Fig.~\ref{fig:accuracy}, our method outperformed the traditional registration method ICP~\cite{besl1992method}, deep feature-based method DCP and DeepGMR, and original PointNetLK. Even with a rotation error threshold less than $0.5^\circ$ and a translation error threshold less than $0.005$, our method can still achieve $0.98$ of success ratio, while ICP has only $0.31$ of success ratio. 
Only DeepGMR and PointNetLK were competitive with our method, but our method still achieved smaller rotation errors. The results clearly indicate that our proposed approach achieved highly accurate alignments.

\begin{figure}[t!]
    \centering
    \includegraphics[width=0.97\linewidth]{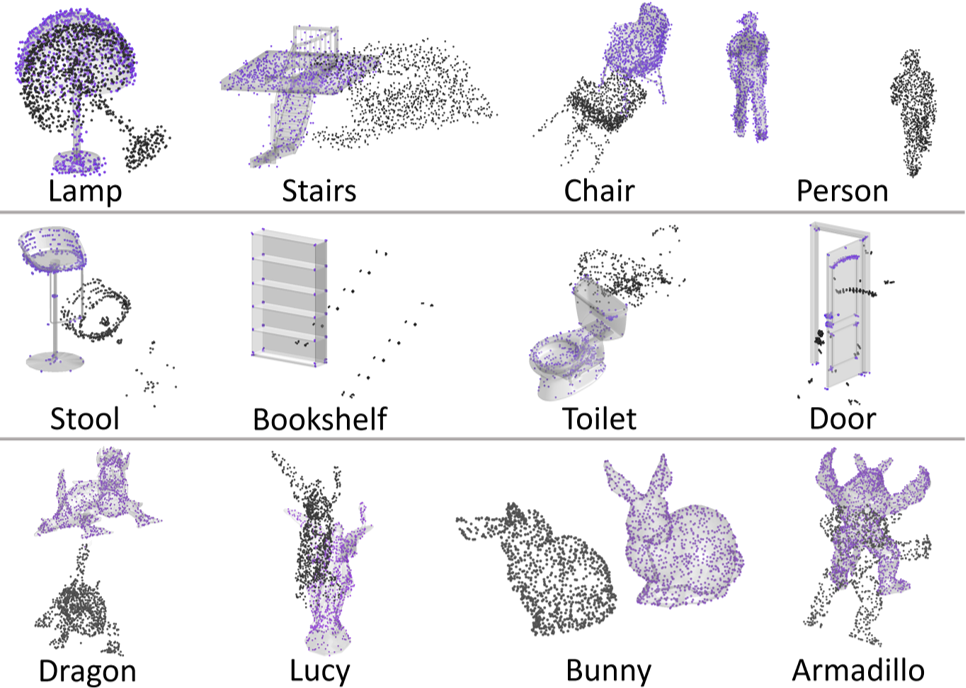}
    \caption[]{\textbf{Generalized registration results.} Visual results of our method on different 3D objects.
    We show the template models as 3D surfaces for better visualization and the source point clouds are in black. The registered point cloud is shown in purple. 
    The 3D models in the first two rows are from ModelNet40, while the last row shows objects from the Stanford 3D scans dataset\protect\footnotemark\addtocounter{footnote}{-1}.}
    \label{fig:object_showcase}
    \vspace{-4mm}
\end{figure}

We also present quantitative results on the unseen object categories in ModelNet40 and ShapeNetCore datasets on different metrics in Table~\ref{tb:accuracy}.
Compared to other methods, our proposed approach achieved extremely low median errors in both rotation and translation. 
This result reveals that our method achieved significant accuracy for most test cases, while only a small portion of them had larger errors. Qualitative results are shown in Fig~\ref{fig:object_showcase}.

\begin{figure}[t!]
\centering
    \begin{subfigure}
        \centering
        \includegraphics[width=0.489\linewidth]{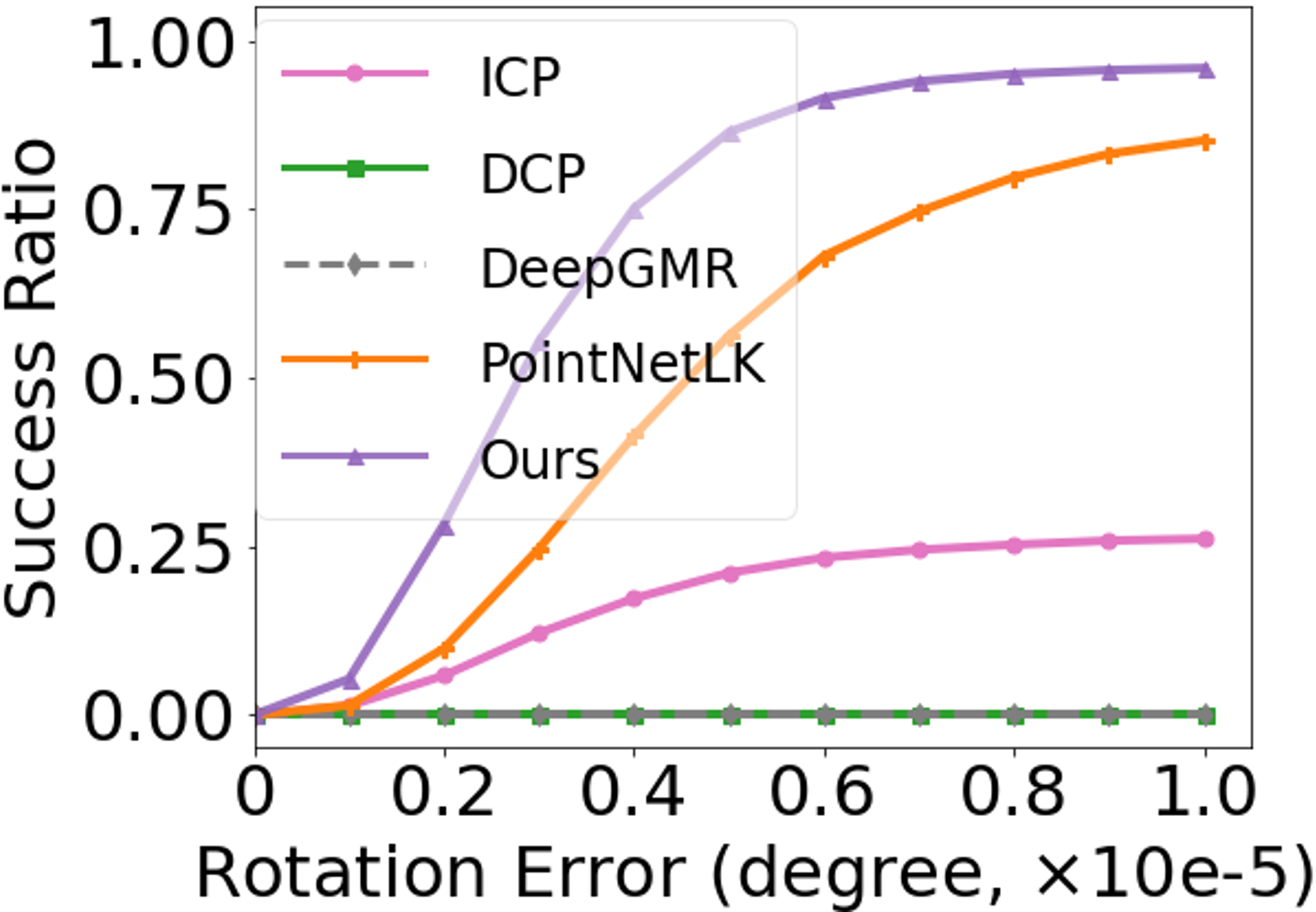}
    \end{subfigure}
    \begin{subfigure}
        \centering
         \includegraphics[width=0.489\linewidth]{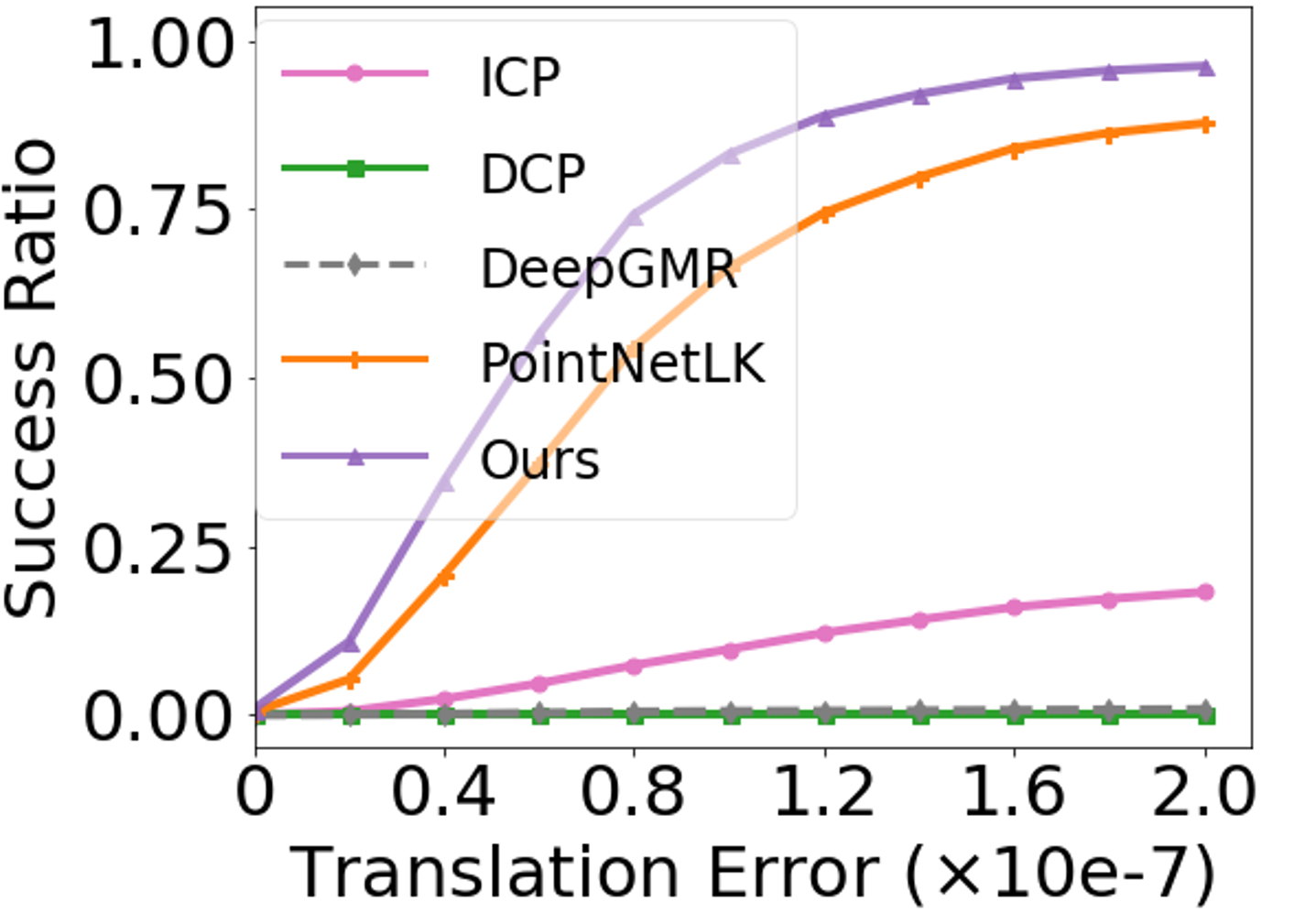}
    \end{subfigure}
    \caption{\textbf{Fidelity analysis.} We set extremely low error thresholds for both rotation and translation during testing. The purple line shows that our method preserved the highest fidelity among other methods. The orange line denotes that the original PointNetLK also achieved reasonable high accuracy. However, with the approximated numerical Jacobian, it lacks fidelity when compared with our method. The green and gray lines indicate the low fidelity of DCP and DeepGMR.}
    \label{fig:fidelity}
    \vspace{-4mm}
\end{figure}

\noindent\textbf{Fidelity analysis:}
\label{ex:fidelity}
We further demonstrate the high fidelity property of our method by setting the maximum rotation error threshold in the range of $[0^\circ, 1\mathrm{e}\text{-}5^\circ]$, and the maximum translation error threshold in $[0, 2\mathrm{e}\text{-}7]$. In Fig.~\ref{fig:fidelity}, we show that under an extremely small fidelity criterion, our approach still achieved higher fidelity than the original PointNetLK and ICP, and also achieved a high success ratio with infinitesimal registration errors. 
The outperformance of our approach attributes to the analytical gradients.

\noindent\textbf{Efficiency:}
\label{ex:efficiency}
Our method is computationally efficient during training and testing. The test time of point correspondence-based methods grows quadratically as the number of points grows. In contrast, our method maintains high efficiency on large-scale data even without the need for graphic processing units. On a single CPU, our method took 0.37 s to process one sample with $10^4$ points in a point cloud, while PointNetLK took 2.87 s, DeepGMR took 2.35 s, and DCP took 222 s.
Please refer to the supplementary material for more results on the efficiency of our method.

\begin{figure}[t!]
    \centering
    \subfigure[Robustness to noise]{
        \includegraphics[width=0.472\linewidth]{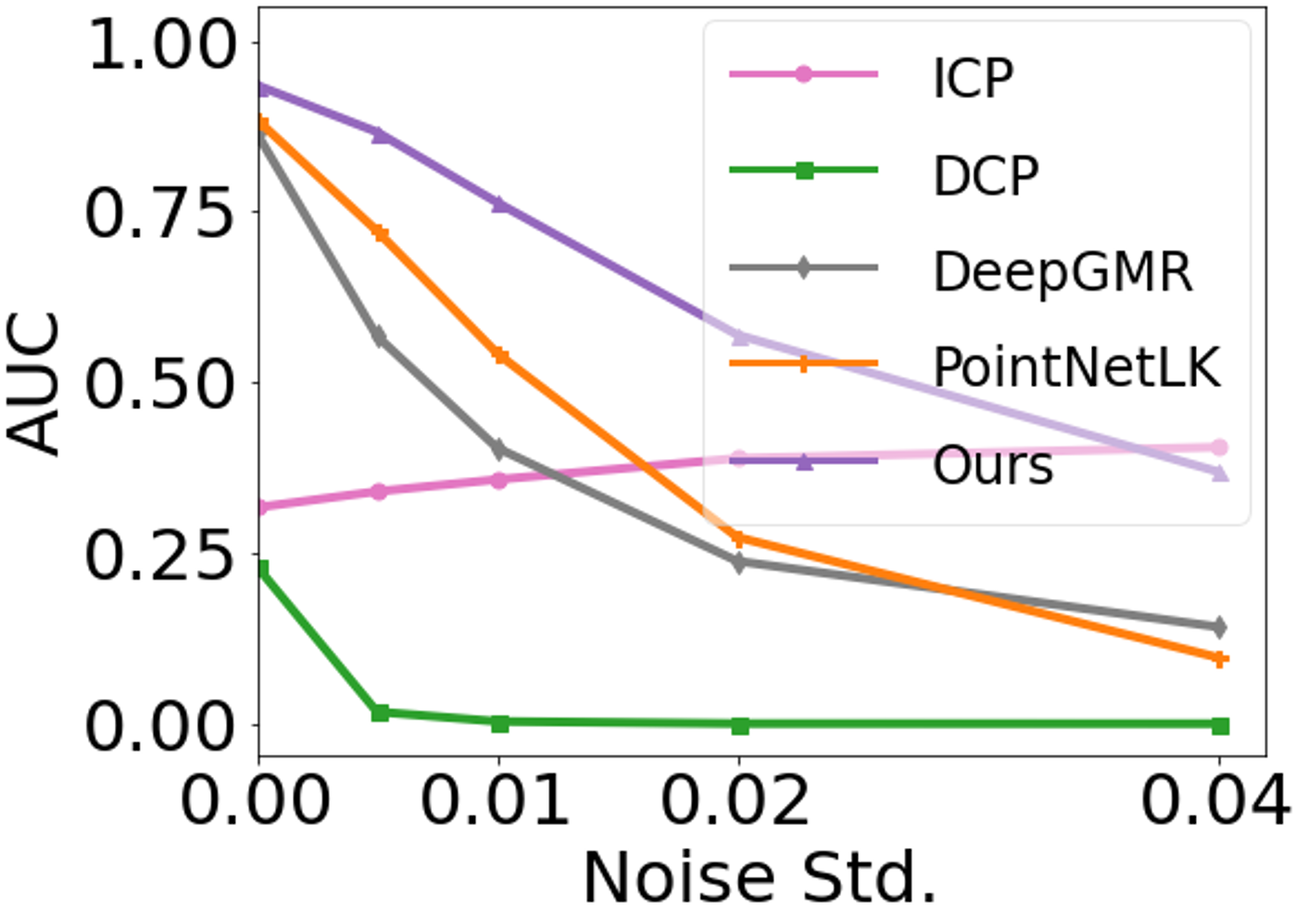}
        \label{fig:noise}}
    \subfigure[Sparse data]{
        \includegraphics[width=0.472\linewidth]{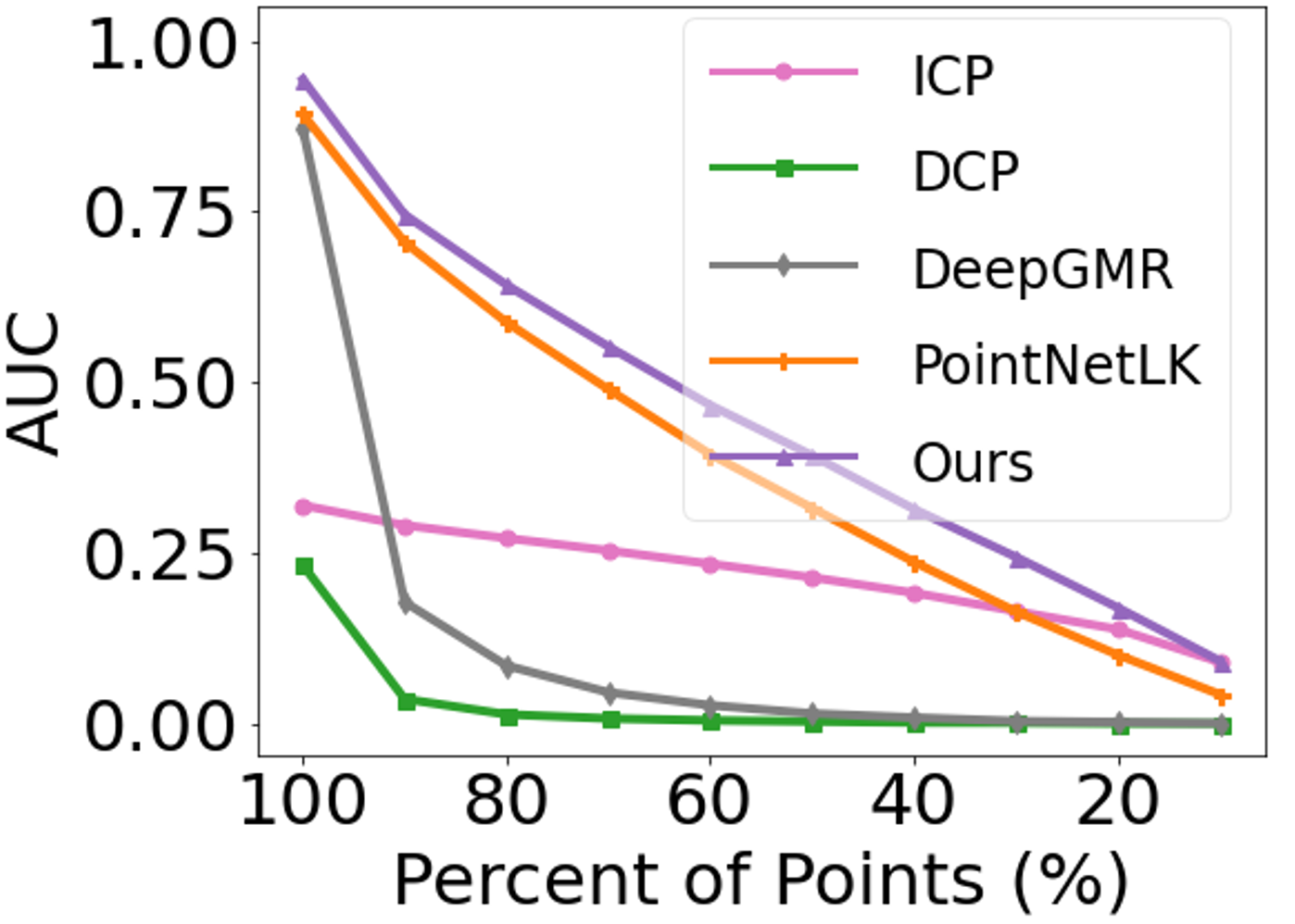}
        \label{fig:sparse}}
    \caption{\textbf{Robustness to noise and sparsity.} 
    In (a), we added Gaussian noise with zero mean and different standard deviations to the source point cloud during testing. Even with relatively large Gaussian noise (std${=}0.04$), our method still achieved around $40\%$ successful registration cases under the success criterion. 
    The numerical PointNetLK had less accuracy on noisy data.
    In (b), we show the registration results with different sparsity levels in the source point cloud. 
    Given $50\%$ of the points from the source point cloud, our method still achieved $0.4$ AUC with the current threshold, while DCP and DeepGMR nearly failed even if $90\%$ of the points were provided.}
    \vspace{-4mm}
\end{figure}

\noindent\textbf{Robustness to noise:}
\label{ex:noise}
To verify our method's robustness to noise, we trained it on noiseless data and then added Gaussian noise independently to each point during the test. Note that we only added noise to the source point cloud, which breaks the one-to-one point correspondences.
Compared with the original PointNetLK and DeepGMR, our approach is more robust to noise even when the source point cloud has large noise (\eg, 0.04), as shown in Fig.~\ref{fig:noise}. DCP failed when large noise was applied.

\noindent\textbf{Sparse data:}
\label{ex:sparse}
In real-world applications, 
point clouds obtained from LiDAR sensors are generally sparse. To test our performance with sparse data, we subsampled the source point cloud at different sparsity levels and evaluated the registration performance against the original dense template point cloud.
Fig.~\ref{fig:sparse} shows that our method maintained a relatively high success ratio than the baselines in these sparse registration scenarios.

\addtocounter{footnote}{+1}\footnotetext{\href{http://graphics.stanford.edu/data/3Dscanrep}{http://graphics.stanford.edu/data/3Dscanrep}.}

\noindent\textbf{Partial data:} 
\label{ex:partial}
We explored the capacity of our method to register point clouds on partially visible data. This captures real-world situations where the data is normally corrupted by occlusions, holes, etc. 
Following similar settings as in~\cite{aoki2019pointnetlk}, we selected both source and template to be partial point clouds by determining which points were visible from certain random camera poses.\footnote{Please refer to the supplementary material for more details.}
When we set the success registration criterion to be a rotation error under $5^\circ$ and translation error under $0.1$, our analytical PointNetLK achieved $0.69$ AUC, while the AUC for ICP, DCP, DeepGMR, and original PointNetLK were $0.21$, $0.22$, $0.35$ and $0.63$, respectively.

\noindent\textbf{Ablation study:}
\label{ex:techniques}
Due to the space limitation, we placed all the ablation studies in the supplementary material to show the impact of different network design strategies and the numerical vs. analytical Jacobian analysis.

\subsection{Performance on real-world 3D scenes}
\label{ex:general_testset}

Here we demonstrate our method's generalizability and accuracy to handle complex real-world scenes obtained from different 3D sensors.

\begin{figure*}[t!]
    \centering
    \includegraphics[width=0.957\linewidth]{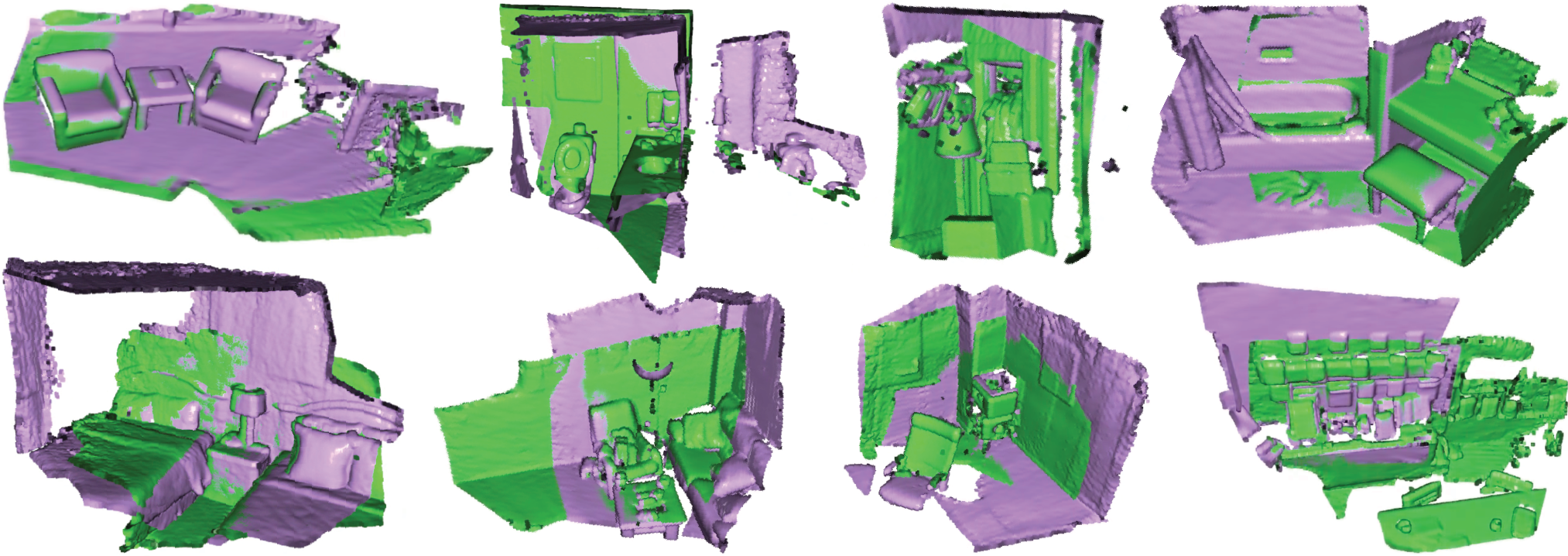}
    \caption{\textbf{Visual results on complex, real-world scenes.} Our voxelized analytical PointNetLK is able to register complex, real-world scenes with high fidelity. These scenes are from 8 different testing categories of the 3DMatch dataset. Purple is the registered scene and green is the template.}
    \label{fig:complex_3dmatch}
\end{figure*}

\noindent\textbf{Generalizability to real-world data:}
\label{ex:complex_scene}
We used the model trained on the ModelNet40 dataset as explained in the experiment setup and directly tested it on the real-world 3D scans from 3DMatch.
In Table~\ref{tb:complex_generalizability}, we show the performance of different methods on 627 indoor 3D scans.

The hybrid learning models DCP and DeepGMR failed when the data had a completely different distribution.
The original PointNetLK lost accuracy while still being relatively robust, given that it generalized to the data that is not well-represented in the training set. 
ICP outperformed other learning-based methods, while only our analytical PointNetLK achieved competitive results.

\noindent\textbf{Voxelization to handle large-scale scenes:}
Table~\ref{tb:complex_generalizability} shows the performance of our proposed voxelized analytical PointNetLK. 
Although we trained our model on a synthetic 3D object dataset, we can still utilize voxelization during testing to drastically improve the registration results of complex scenes compared to the vanilla analytical PointNetLK. 
We also show that with more voxels to discretize the 3D space, or with more points included in each voxel, we can further improve the registration performance.
As we stated in Section.~\ref{sc:voxelization}, without voxelization, the PointNet feature function is not able to capture large-scale point cloud features in a global feature vector. While with the partitioning of the 3D space, we were able to include more geometric information in each local feature descriptor.
Fig.~\ref{fig:complex_3dmatch} shows qualitative results for our voxelized analytical PointNetLK.

\begin{table}[t!]
\begin{adjustbox}{width=\columnwidth}
\centering
\begin{tabular}{lcccc}\toprule
                 & \multicolumn{2}{c}{Rot. Error (degrees)} & \multicolumn{2}{c}{Trans. Error (m)} \\ \cmidrule(lr){2-3} \cmidrule(lr){4-5}
                Algorithm & RMSE~$\downarrow$ & Median~$\downarrow$ & RMSE~$\downarrow$ & Median~$\downarrow$ \\ \midrule
ICP~\cite{besl1992method}   & 24.772 & \textbf{4.501} & 1.064 & \textbf{0.149} \\
DCP~\cite{wang2019deep}   & 53.905 & 23.659 & 1.823 & 0.784 \\
DeepGMR~\cite{yuan2020deepgmr} & 32.729 & 16.548 & 2.112 & 0.764 \\
PointNetLK~\cite{aoki2019pointnetlk} & 28.894 & 7.596 & 1.098 & 0.260\\
Ours (no voxelization)  & \textbf{15.996} & 4.784 & \textbf{0.738} & 0.169 \\  \midrule
Ours (27 voxels, 37 points) & 10.068 & 3.623 & 0.427 & 0.127 \\
Ours (8 voxels, 125 points) & 7.666 & 3.362 & 0.405 & 0.123 \\
Ours (27 voxels, 148 points) & 7.999 & 2.714 & \textbf{0.308} & 0.101 \\
Ours (8 voxels, 500 points) & \textbf{7.165} & 2.799 & 0.417 & 0.105 \\
Ours (8 voxels, 1,000 points) & 7.656 & \textbf{2.535} & 0.355 & \textbf{0.096} \\
\bottomrule       
\end{tabular}
\end{adjustbox}
\vspace{0.01ex}
\caption{\textbf{Performance on complex, real-world scenes.} All learning methods were trained on the synthetic ModelNet40 dataset and tested on the real-world 3DMatch dataset to investigate their generalizability across data with different distributions. Our proposed voxelized analytical PointNetLK generalized to real-world, complex, unseen scenes despite being trained on synthetic objects. State-of-the-art hybrid learning models had poor generalizability. The classical ICP achieved average performance but still outperformed the learning methods. The number of voxels means how many voxels in the 3D space, and the number of points means the maximum number of points in each voxel.}
\label{tb:complex_generalizability}
\vspace{-4mm}
\end{table}

\begin{figure}[t!]
\centering
    \begin{subfigure}
        \centering
            \includegraphics[width=0.489\linewidth]{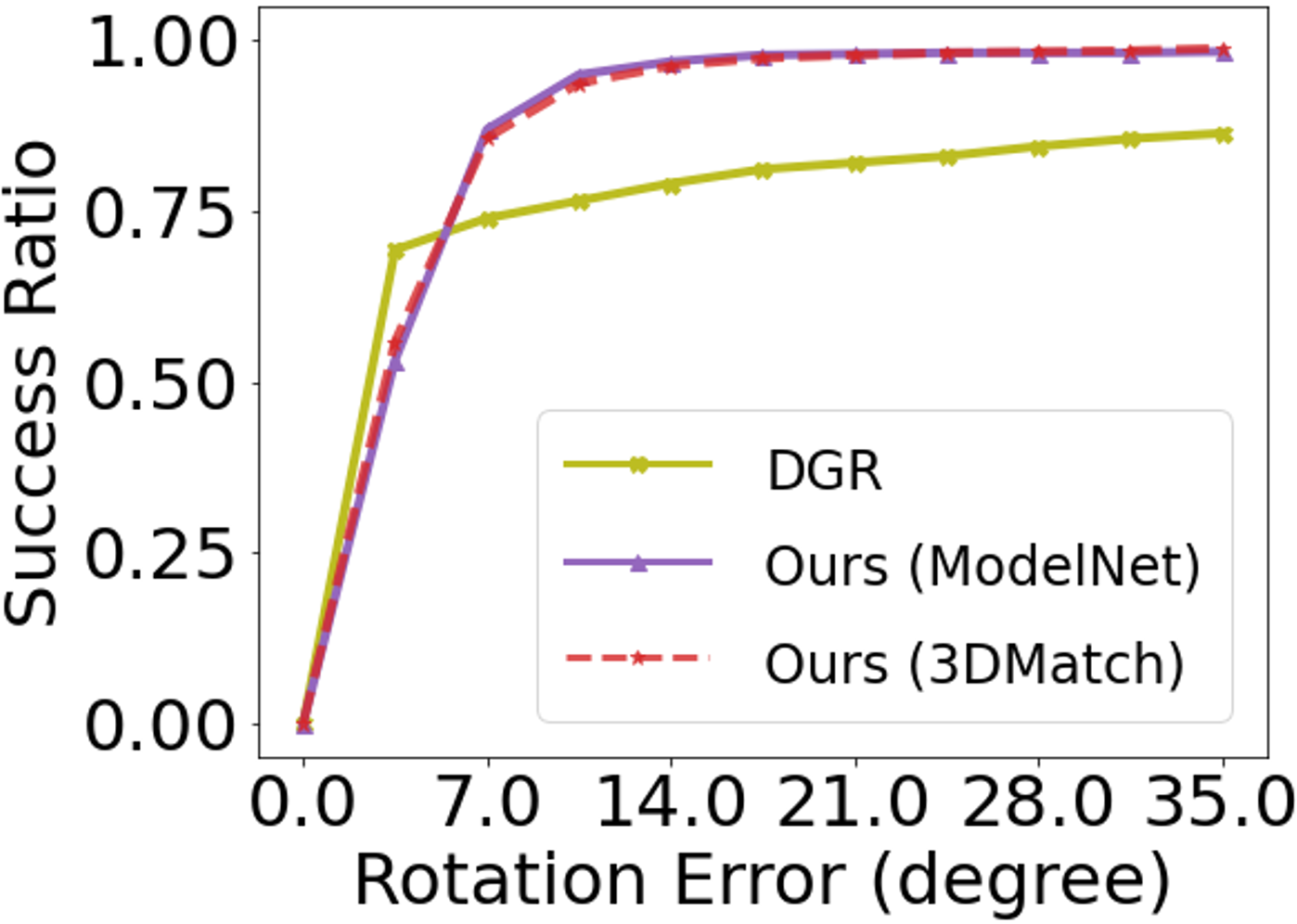}
    \end{subfigure}
    \begin{subfigure}
        \centering
         \includegraphics[width=0.489\linewidth]{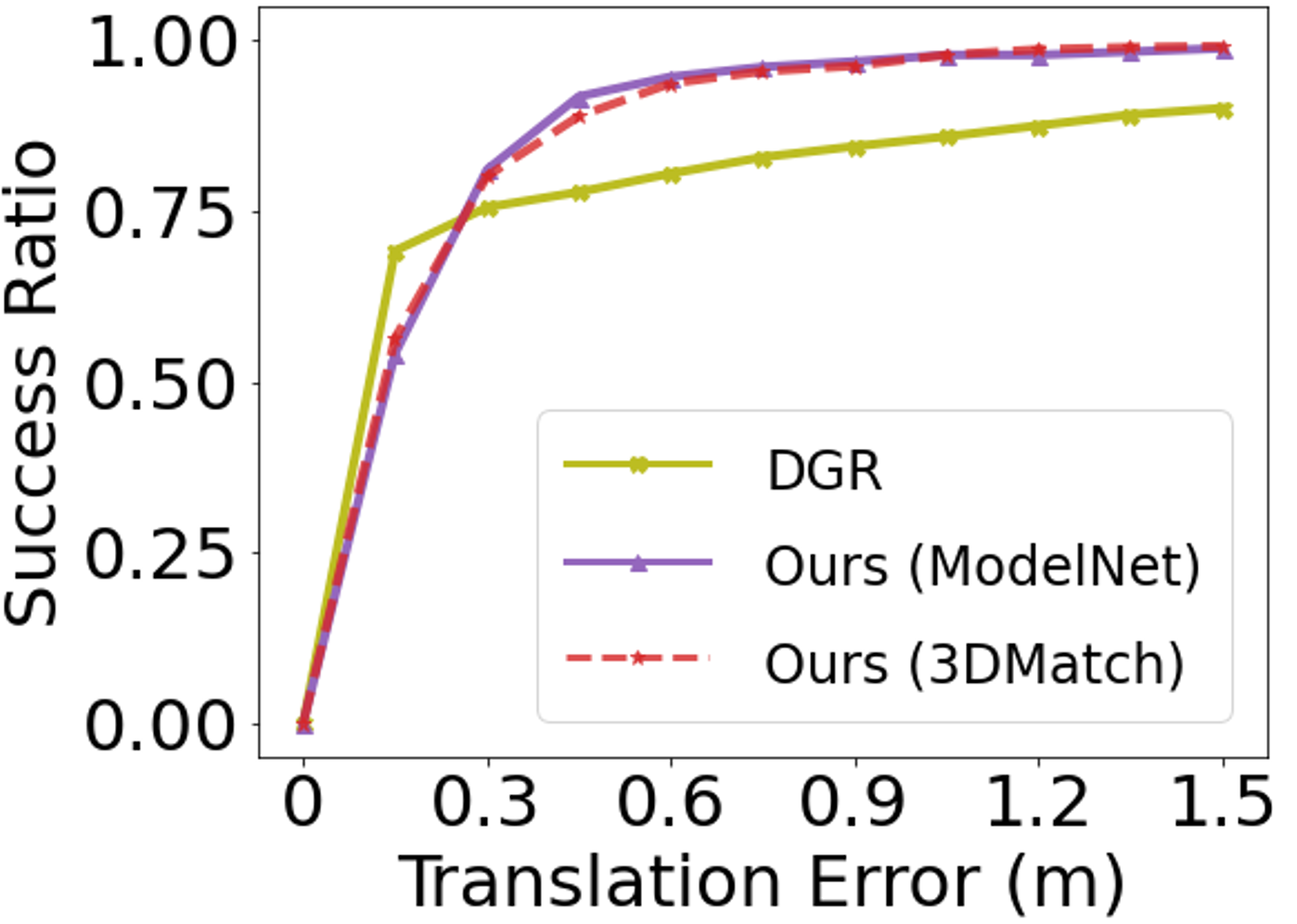}
    \end{subfigure}
    \caption{\textbf{Comparison with DGR~\cite{choy2020deep}.} 
    We report the success ratio of our method and DGR for rotation and translation errors. Note that our method, no matter trained on ModelNet or 3DMatch, achieved better performance where more successful registration cases were observed. For rotation errors larger than $5^{\circ}$ and translations larger than $0.2$ m, our method still had successful registration cases while DGR had poor performance.
    Note that training our model on the 3DMatch dataset will likely not improve the performance, highlighting the generalizability of our analytical PointNetLK on different 3D sensors.} 
    \label{fig:compare_dgr}
    \vspace{-4mm}
\end{figure}

\noindent\textbf{Comparison with DGR:}
The Deep Global Registration (DGR)~\cite{choy2020deep} is a state-of-the-art hybrid learning registration method on complex real-world 3D scenes. 
Aiming for a fair comparison, we trained both our analytical PointNetLK and DGR on the 3DMatch dataset with an initial random perturbation in the range of $[0^\circ, 45^\circ]$ in rotations and $[0, 0.8]m$ in translations. 
We randomly sampled 1,000 points for our model and used all the points for DGR since it relies on a pre-trained FCGF~\cite{choy2019fully} model.
Our proposed method was competitive with 
DGR on 3DMatch when trained on either ModelNet40 or 3DMatch dataset, as shown in Fig.~\ref{fig:compare_dgr}. The results demonstrated the superior generalization ability of our method. Since it does not require similar distribution between training and testing set, our method is more likely to be robust in unseen data with different distributions.

\section{Conclusion}
\label{discussion}
We revisited PointNetLK and proposed a novel derivation using analytical Jacobians to circumvent the numerical instabilities of the original PointNetLK.
As a deep feature-based registration method, our method leverages deep point feature representation and the intrinsic generalizability of the classic IC-LK. This allows our method to generalize to different object categories and data obtained from different sensors.
To manage complex, real-world scenes, we took advantage of our decomposable formulation and proposed a voxelization strategy that is only applied during testing.
Our experiments showed that our method has high fidelity and robustness for point cloud registration on synthetic data and generalizes to complex, real-world scenes.

\noindent \textbf{Acknowledgments}:
The authors would like to thank Ioannis Gkioulekas, Ming-Fang Chang, Haosen Xing, and the reviewers for their invaluable suggestions that significantly improved this work. Also, we thank Ming-Fang Chang for the assistance with the related work section, Chen-Hsuan Lin for paper organization suggestions, and Jianqiao Zheng, Haosen Xing, and Chaoyang Wang for the fruitful discussions throughout the project. This work was supported by Argo AI and the CMU Argo AI Center for Autonomous Vehicle Research.

{\small
\bibliographystyle{ieee_fullname}
\bibliography{egbib}
}

\end{document}